\documentclass[11pt]{article}
\usepackage{graphicx}

\title{\bf Textual Fingerprinting with texts from\\ Parkin, Bassewitz, and Leander}
{\author{\small Christoph Schommer \\ 
              \small University of Luxembourg\\ 
              \small Dept. of Computer Science - ILIAS Laboratory\\
              \small 6, Rue Richard Coudenhove-Kalergi, 1359 Luxembourg, Luxembourg\\
              \small Email: christoph.schommer @ uni.lu
\and
             \small Conny Uhde\\
             \small JW Goethe-University Frankfurt am Main\\
             \small Dept. of Computer Science and Mathematics\\
             \small Robert-Mayer-Str. 11-15, D-60486 Frankfurt am Main, Germany.\\
             \small Email: uhde@cs.uni-frankfurt.de}
\date{\today}
\begin{document}

\maketitle

\begin{abstract}
Current research in author profiling to discover a legal author's fingerprint does not only follow examinations based on statistical parameters only but include more and more dynamic methods that can learn and that react adaptable to the specific behavior of an author. But the question on how to appropriately represent a text is still one of the fundamental tasks, and the problem of which attribute should be used to fingerprint the author's style is still not exactly defined. In this work, we focus on linguistic selection of attributes to fingerprint the style of the authors Parkin, Bassewitz and Leander. We use texts of the genre \textit{Fairy Tale} as it has a clear style and texts of a shorter size with a straightforward story-line and a simple language.
\end{abstract}

% \twocolumn

% ==========================================================
\section{What is it about?}\label{secWhatIsItAbout}
% ==========================================================

The\footnote{This work has been supported by the University of Luxembourg within the project \textit{TRIAS} -  Logic of \textit{T}rust and \textit{R}eliability for \textit{I}nformation \textit{A}gents in \textit{S}cience}. forensic linguistics is concerned with a verification process for the decryption of texts and the analysis through pattern discovery. In this respect, verification means the usage of existing and well-known stylistic attributes to discover an individual (linguistic) fingerprint. However, it is still a controversial discussion, if such a linguistic fingerprint is a clear indication per se: stylistic tests assume that typical attributes are directly influenceable by the author and that a certain number of attributes still keep constantly, even though the author changes consciously the behavior or the own style \cite{DM93}. However, following Dixon and Mannion to their evaluations to the texts of Oliver Goldsmith, it may be observed that an appropriate selection of stylistic attributes take a risk: Goldsmith's style is characterized by an adaptive fluency, where he adapts his onw style in a reported speech of the respective actor. To identify Goldsmith's characteristic style attributes, Dixon and Mannion compared his essays with those of four contemporary writers. They found out that two of the four writers show a suspicious similarity with Goldsmith as they originate from the same irish area, living in the english exile \cite{MD97}.

Additionally, stylistic attributes that are influenced by the genre may interfere the individual style \cite{BVHT96}. This leads to the conclusion that only texts of the same domain are affiliated with each other. Using the texts of the Nijmegen-corpus, Baayen et al. have analyzed the differences of diverse authors of the same genre as well as the texts of authors who represent different genres: they have found out that texts of the same genre are generally more similar than texts of the different genre that are from the same author.

% --------------------------------------------------------------------------------
\section{Style Discovery}
% --------------------------------------------------------------------------------
Stylometry refers to the measurement of the style with the aim to fingerprint a text following a certain number of linguistic attributes, to conclude the authorship of a text and/or to order texts following their chronology \cite{OAK98}. The content, the meaning and the correctness of the text is not concerned. The general ambition is to discover those attributes that difference texts sufficiently \cite{MQ78}. Generally, the data is analyzed statistically taking numerical attributes into account but disregarding categorical attributes. Figure of speeches like metaphor and symbols are clear defined indeed, but are not to be discovered automatically. In \cite{OAK98}, Oakes writes that any linguistic occurrence can be taken for the stylometric analyis as the attribute can be expressed by a numerical attribute. However, it must be assured that the attribute is relevant for other genres as well \cite{MO00}. Another important aspect is the differentiation of linguistic attributes of whether they are consciously controlled by the author or not \cite{LED90}. Many examinations take explicitly unconscious stylistic attributes as the relevant discrimination criterion as they are a stronger sign of a stylistic fingerprint. However, this includes the existence of stylistic attributes that stay constantly through the whole text and the existence of linguistic attributes that adapt \cite{LAA95}. In this respect, we focus on a differentiation of conscious and unconscious stylistic attributes, well noting that diverse authors differ more in their style than texts of an individual author. Furthermore, texts of an individual authors differ more than passages within a text \cite{DIM04}. We therefore conclude that an appropriate consistence of a continuous usage of conscious and unconscious stylistic attributes must be generally secured. Very generally, linguistic attributes refer either to a statistic frequency or to the differentiation of the vocabulary \cite{SFK01} - under the assumption that author differ in their vocabulary and that they control their vocabulary rather limited than specific. The vocabulary is then queried by habit, it is performed automatically and therefore constant, appropriate for text classification \cite{HO04}.

Several examinations have shown that a few stylistic attributes are insufficient for the differentiation of authors as they produce \textit{pairs of authors} classifying in the same category. \cite{DIM04} and \cite{RUD98} suggest a wider spectrum of attributes leading to a better success and argue that the selected attributes can be ordered depending on their significance in respect to a classification the genre. 

The research on a stylistic analysis for author profiling has been started years ago, when Mendenhall \cite{T87} and Mascol \cite{C88} examined literary verses of the New Testament by considering aspects like frequency of words the length of sentences. They assumed that authors produce different texts, with different style and features. Many statistical examinations followed, for example to discover text features that may appear constantly. Many attributes have been found and mathematical issues proven, for example the Yules characteristics, Zipf's law and the Hapax Legomena. \cite{MF64} has shown that a statistical relevance on a low number of textual data can be expressed and computed by a Bayesian statistics, which makes it applicable for a contribution to author profiling. However, to conclude that a text is written by a specific author has not often clear and misclassifications has been done.

Since that time, many other attributes have been examined, for example the number of words of a certain wordclass \cite{AS98}, \cite{FRS96}, syntax analysis \cite{BH96}, \cite{SE00} and word phrases \cite{F89}, and grammatical failures \cite{KM03a}. Many examinations combine some of these attributes as well as different methods from statistics and machine learning, for example principal component analysis, support vector machines, and cluster analysis \cite{D00}.

% ========================================================
\section{About the evaluation environment}
% ========================================================

In this work, we understand \textit{Author Profiling} as a way to identify authors by a certain number of linguistic (numerical) attributes and to assign texts correctly to them. In this respect, our hypothesis is that - based on the assumption that there exists a potential style identification - an stylistic identification of authors can be done with quality if we can find a sufficient number of expressive attributes, which describe the author's behavior in respect of characteristic and dissimilarity; and that allows an application of machine learning methods for an demonstrative evaluation. Nevertheless, to perform an empirical study in order to discover the author's style is mostly characterized by a linguistic detail, namely the principal use of attributes that are applicable within a computer-based analysis. And an independence of these attributes must be adjusted as well.

\subsection{The genre we use}
We focus on several texts from the genre \textit{Fairy Tale}. The texts are in german language. Fairy tales have been selected as they are per se an excellent differentiator to other texts; they are distinguished by a clear style and author-independent. Mostly, fairy tales are of shorter size, they are amusing stories with fantastic content without a reference of time. The storyline is straightforward, the language simple.

Although common speech texts are more difficult to differentiate as they do not have to follow a certain style per se, they are contradictory to the texts of a technical texts serving the presentation and the critical discussion with specific contextual aspects. Common speech texts are non-coded and of daily use, easy understandable, but less syntactically defined. Technical texts are often related to science and therefore underlie certain criterions. The understanding of a technical text is highly depending on the style. Nevertheless, authors try to keep their individual style, taking into account the correct use of orthography, syntax and punctuation. The text is often impersonal and written in present tense. We have therefore started our examinations with a comparison between selected texts from \textit{Fairy Tale} and \textit{Common Speech} and \textit{Technical Language}, respectively. Approximately 10 authors per each genre with 3-5 documents have been selected. 

% ------------------------------------------------------------------------------------------------------
\subsection{Attribute Selection}
In concern of the attributes, we take into account linguistic attributes as much as they significantly contribute to the author's style, but filter those out that are dependent from another. For the evaluations, we have used more than 30 attributes from statistics or linguistics, for example

\begin{itemize}
     \item \textit{Number of Words and Number of distinct words}, where punctuation marks are disregarded. 
     \item \textit{Frequency of personal pronoun}. Depending on the genre, the personal pronoun is assessed; for example, the word \textit{I} receives a higher weight in scientific texts than in fairy tales since we may assume that scientific texts follow a more neutral description (passive) or uses the \textit{We} instead.
     \item \textit{The word with the highest frequency}. 
     \item \textit{Word length in average}.
     \item \textit{Record length}. We use this attribute although \cite{SMI83} mentions that the record length is not expressive and applicable as a single attribute. The disadvantage is the author's control and capability to imitate, especially against the punctuation. This makes it less suitable to older texts. \cite{TAL72} agree that the attribute record length is a weak measurement for the author's style but is useful when focusing on their distributions.
     \item \textit{Yules characteristic value k}. This value bases on the assumption that the occurrence of a word is random and underlies the distribution of Poisson. The more the words are repeated, the higher k.
     \item \textit{Hapax Legomena}, the number of words that occur exactly once in the text. This is to measure the author's disposition to use or to avoid synonyms.
     \item \textit{Sentence Structure}. This attribute describes the author's disposition to prefer main clauses or subordinate clauses. We measure this by the percentage of hypotaxis in the text.
     \item \textit{Value of the the type-token proportion}. Let \textit{n} the number of tokens (words) in the text and \textit{v} the number of different tokens, then the type-token proportion \textit{r} is the fraction between \textit{v} and \textit{n}.
     \item The \textit{Entropy} of the text. The length of each text source is set to a fixed number of words.
\end{itemize}

Furthermore, we have concerned with \textit{stop words} and calculated (per text) the number of words occurring exactly once, the stop word itself that occurs most frequently, and its frequency and percentage. Using the thesaurus of the University of Leipzig\footnote{see Wortschatz - http://wortschatz.uni-leipzig.de/}, we additionally calculate the frequency class, which refers to how often a token occurs in comparison to any occurrence.

\begin{figure}[h]
   \centering
   \includegraphics[width=9.5cm]{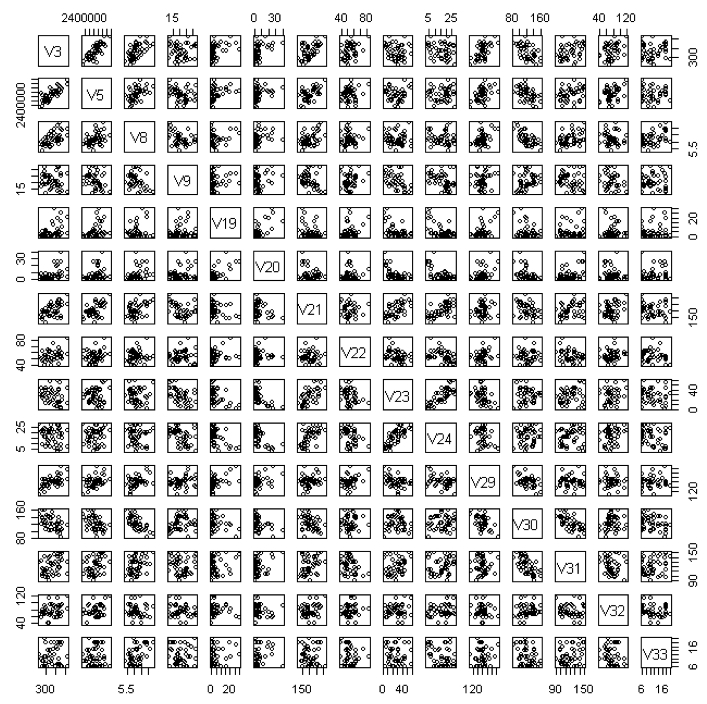} 
   \caption{Received list of independent attributes for the genre \textit{Fairy Tale}, plotted in a quadratic and symmetric comparison matrix.}
   \label{fig:fingerprintA}
\end{figure}

% ------------------------------------------------------------------------------------------------------
\subsection{Attribute Filtering}
A problem in using the selected attributes is that some of them may dependent from others. Furthermore, this depends on the genre very strict, so that attributes of a genre must be preprocessed individually. We therefore have used the statistical method of \textit{Plots} to pairwise visualize the attribute's behavior. For example, the attributes \textit{Entropy}, \textit{Type-Token Ratio}, and \textit{Average word frequency} are dependent whereas \textit{Number of Hapax Legomena} and \textit{Yules characteristics} are not. Figure \ref{fig:fingerprintA} shows the preprocessed filtering result of texts from the genre \textit{Fairy Tale}, i.e., the pairwise distribution of independent attributes. The presented plot is symmetric. 

% ==========================================================
\section{Selected Fingerprinting Results}\label{secResults}
% ==========================================================
We have enriched this calculation with diverse statistical methods like principal component analysis or bivariate statistics to visualize and calculate the most interesting and reliable attributes and have applied machine learning in a different way through demographic clustering or a genetic algorithm. Some evaluations have been done with the \textit{IBM Intelligent Miner V8}, some with the statistical program  \textit{R}, and  \textit{ClusGen}: this a self-programmed genetic simulation to classify the whole text corpus to classes. The idea focus on the assumption that a representative \textit{median vector} for a collection of texts - that come from the same author - dynamically exists; and that the texts of one author are representative enough for all texts of the corresponding author. We have initiated the process of fingerprinting while interpreting all discovered results. 

% ------------------------------------------------------------------------------------------------------
\subsection{\textit{Fairy Tale} against \textit{Common Speech} and \textit{Technical Language}}
\begin{figure}[htbp]
   \centering
   \includegraphics[width=9.5cm]{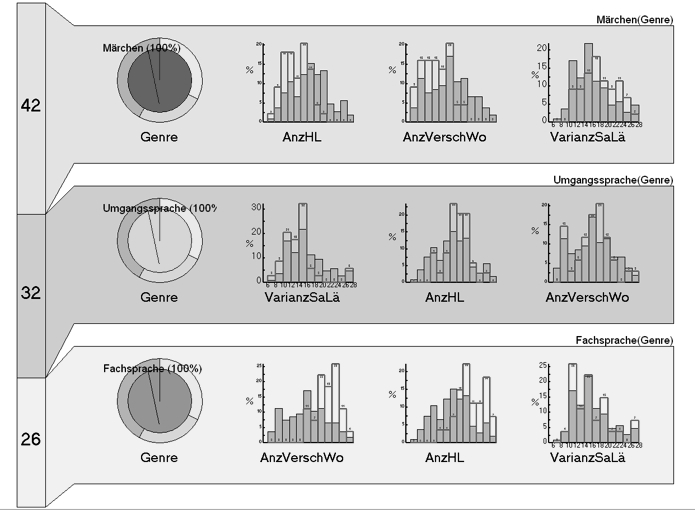} 
   \caption{Bivariate Statistics with three variables against the genre (\textit{Fairy Tale} (top), \textit{Common Speech} (middle), and \textit{Technical Language} (bottom)).}
   \label{fig:fingerprintB}
\end{figure}

% ------------------------------------------------------------------------------------------------------
To get an overview of each genre per se and to characterize these texts as well, Figure \ref{fig:fingerprintB} shows the result of a bivariate statistics against the attributes \textit{Genre}. We observe that 42\% of the text set are from \textit{Fairy Tale}, 32\% from \textit{Common Speech}, and 26\% from \textit{Technical Language}. The variables are ordered following their chi-square values, meaning that the discrepancy between the distribution of the corresponding attribute inside a genre (inner ring for \textit{Genre}, non-colorized distribution of the other attributes) to the whole text population (outer ring for \textit{Genre}, colorized distribution of the other attributes) represents the significance and therefore position of an attribute in each region: the more different the distribution the more it is positioned to the left.

In this respect, \textit{AnzHL} represents the most important attribute to \textit{Fairy Tale} (M\"archen) and is more significant to its region than in any other genre. Furthermore,  the attribute \textit{Number of Hapax Legomena (AnzHL)} is characterized by a distribution of high values in \textit{Technical Language}, but is low distributed in \textit{Fairy Tale}. On the other side, the attribute \textit{Number of different tokens (AnzVerschWo)} has a distribution with high values in \textit{Technical Language} but with lower values in \textit{Common Language}. Following this, we can assume that in \textit{Fairy Tale} the words occur more seldom only once (\textit{AnzHL}); the number of different words (\textit{AnzVerschWo}) is rather low while having longer sentences (\textit{VarianzSaL\"a}). Texts from \textit{Technical Language}, however, use a more extended vocabulary (\textit{AnzVerschWo}), where words occur more often only once (\textit{AnzHL}).

With text clustering, we have observed several clusters representing only texts of an individual genre. For \textit{Fairy Tale}, relatively many attributes have a distribution that is surprisingly higher than in all texts, for example the number of adjectives (\textit{AnzAdjektive}), the number of Hapax Legomena (\textit{AnzHL}) and the Yules characteristics (\textit{YulesK}); on the other side, the number of verbs (\textit{AnzVerben}) and the frequency class (\textit{H\"aufKla}) are quite low distributed. Five authors share this cluster, the most used words are \textit{ich} and \textit{und}. The number of parataxis are relatively lower, the number of hypotaxis relatively higher. We may conclude that the general style is descriptive and figurative, because many adjectives and synonyms are used. It is explainable, since longer sentences exist that are often nested.

Generally, the tests have shown a controversial face. Within the selected texts, the style of only some authors have been constantly, meaning that there exist a set of attributes being equally distributed. This is for the authors \textit{Parkin} and \textit{Bassewitz} as their texts have been selected within a textbook. On the other side, the texts of \textit{Leander} are widespread, although they are also taken from an individual text collection.

% --
\subsection{Parkin's Style}
\begin{figure}[htbp]
   \centering
   \includegraphics[width=9.5cm]{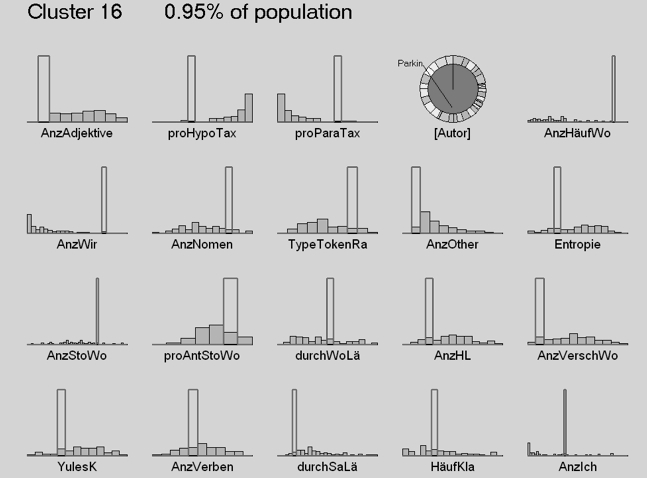} 
   \caption{The Parkin-Cluster, Genre \textit{Fairy Tale}.}
   \label{fig:fingerprintC}
\end{figure}

% --
Figure \ref{fig:fingerprintC} shows a cluster that only bases on texts from the author \textit{Parkin}. We observe a low number of adjectives (\textit{AnzAdjektive}) in all the clustered texts, a high number of parataxis  (\textit{proParaTax}), and a high number of  \textit{we}'s (\textit{AnzWir}). Parkin's style is further characterized by a high type-token ratio (\textit{TypeTokenRatio}), a low number of different words (\textit{AnzVerschWo}), and a low number of Hapax Legomena  (\textit{AnzHL}). He prefers parataxis, the averaged length of sentences is low as well as the number of different words: he therefore tends to repeat words, favors nouns but not adjectives at all.

% --
\subsection{Bassewitz's Style}
\begin{figure}[htbp]
   \centering
   \includegraphics[width=9.5cm]{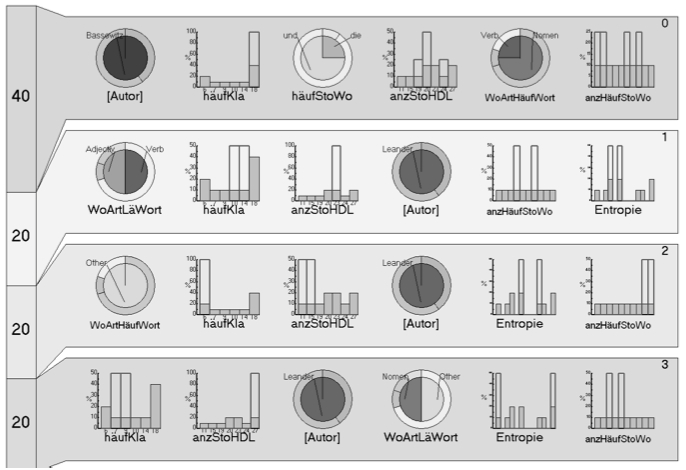} 
   \caption{The Bassewitz-Cluster and the Leander-Clusters, Genre \textit{Fairy Tale}.}
   \label{fig:fingerprintD}
\end{figure}

% --
The Figure \ref{fig:fingerprintD} shows a snapshot of a demographic clustering result that bases on 
\begin{itemize}
    \item four texts of the authors Bassewitz, taken from \textit{Peterchen's Mondfahrt}
    \item six texts of the author Leander, taken from \textit{Tr\"aumereien an franz\"osischen Kaminen}
\end{itemize}
within the genre \textit{Fairy Tale}. Bassewitz's texts are characterized by a high-valued frequency class (\textit{H\"aufKla}) and a disproportionately high occurrence of verbs. Additionally, the word type of the longest word belongs almost to the same class as well as the word type of the most frequently word, the most frequently stop word, and the number of Hapax Legomena.

\subsection{Leander's Style}
% --
The texts of Leander are more mixed, not aiming at an uniform style. We observe that Leander's texts are more different than those from Bassewitz - although both are taken from textbook. Taking the genetic simulation program, we have received classification rates by an adaptive calculation of a new median vector. With this, the classification rates are above 90\% for the whole test corpus. Inside the genres, the classification rate for \textit{Fairy tail} has been around 93\% but for \textit{Common speech} and \textit{Technical language} nearly 100\%.

% ---------------------------------------------------------------------------------------------
\section{Conclusions}\label{secConclusions}
% ---------------------------------------------------------------------------------------------
At first glance, it seems not even conceivable that an author's style or even more, a fingerprint, can be discovered: the number of evaluated texts is quite low and a legal forecast therefore not feasible. But the evaluations prove that through the selection of linguistic attributes, an author can be described within his texts; and even more, the usage of texts within an author's textbook like Bassewitz's \textit{Peterchen's Mondfahrt} may certainly have its eligibility to characterize his style and to allow to score texts in this microcosm. For the authors \textit{Parkin} and \textit{Bassewitz} this is be observed as Parkin's style is documented by a couple of attributes sharing an individual distributive behavior, whereas Bassewitz uses - for example - throughout words that occur seldom. 

To extend these tests to a larger text corpus, to enrich the given set of attributes by \textit{subjective} linguistic attributes, and to generalize our results to other text corpora will be one of our next responsibilities. In this respect, we understand a \textit{subjective linguistic attribute} as a personal statement of the author himself, like for example \textit{I believe that} or \textit{I certainly agree} to express the personal beliefs and intentions. Furthermore, open questions arise like how representative these results are or how appropriate a scoring engine - that assigns for example a text of \textit{Parkin} correctly - is not tested yet.  Our next steps will concern these questions, we also follow up on further examinations. Generally, we strongly believe in this way of style analysis and author recognition, and hope to discover attributes that uniformly relies on our hypothesis.


{\small
\begin{thebibliography}{4}
\bibitem{AS98} G. Avneri, S. Argamon, M. Koppel: Routing documents according to their style. Intl. Workshop on Innovative Internet Information Systems, 1998.
\bibitem{BH96} T. F. Baayen, H. v. Halteren: Outside the cave of shadows: using syntactic annotation to enhance authorship attribution. Literary and Linguistic Computing, (3):121-130, 1996.
\bibitem{BVHT96} H. Baayen, H. von Halteren, F. Tweedie: Outside the cave of shadows: using syntactic annotation to enhance authorship attribution. Literary and Linguistic Computing, (3):121-130, 1996.
\bibitem{C88} C. Maskol: Curves of pauline and pseudo-pauline style i+ii. Unitarian Review 30:452460, 1988.
\bibitem{D00} D. Khmelev: Distributed authorship resolution using relative entropy for Markov Chain of letters in texts. 4th Intl Conference on Quantitative Linguistics Association, 2000.
\bibitem{DIM04} F. Dimpel: Computergest\"utzte textstatistische Untersuchungen an mittelhochdeutschen Texten. T\"ubingen: Francke, 2004.
\bibitem{DM93} P. Dixon, D. Mannion: Goldsmith's Periodical Essays: A Statistical Analysis of Eleven Doubtful Cases. Literary and Linguistic Computing, 8:1 19, 1993.
\bibitem{F89}: F. Smadja: Tge missing link. Journal of the Association for Literary and Linguistic Computing, 4(3), 1989.
\bibitem{FRS96}D. Holmes, R. S. Forsyth: Features finding for text classification. Literary and Linguistics Computing, 11(4):163-174, 1996.
\bibitem{HO04} D. L. Hoover: Another Perspective on Vocabulary Richness. Journal on \textit{Computers and the Humanities}, Springer, pp. 151-178, 2004.
\bibitem{KM03a} J. Schler, M. Koppel: Exploiting stylistic idiosyncrasies for authorship attribution. In Proceedings of IJCAI'03 Workshop on Computational Approaches to Style Analysis and Synthesis, 2003.
\bibitem{LAA95} N. M. Laan: Stylometry and Method. The Case of Euripides. Oxford Journals, Literary and Linguistic Computing, pp. 271-278, 1995.
\bibitem{LED90} G. Ledger: Re-Counting Plato: A Computer Analysis of Plato's Style. Clarendon Press. 1990.
\bibitem{MD97} D. Mannion, P. Dixon: Authorship Attribution: the Case of Oliver Goldsmith. Journal of the Royal Statistical Society (Series D): The Statistician, 46:1-18, 1997.
\bibitem{MF64} D. L. Wallace, F. Mosteller: Applied Bayesian and classical inference. Springer, 1984. 
\bibitem{MO00} T. Mcenery, M. Oakes: Authorship Identification and Computational Stylometry. Handbook of Natural Language Processing. pp. 545-562, 2000.
\bibitem{MQ78} M. A. Queen: Literary Detection. How to prove Authorship and Fraud in Literature and Documents. New York, 1978.
\bibitem{OAK98} M. P. Oakes: Statistics for Corpus Linguistics. Edinburgh Textbooks in Empirical Linguistics. Edinburgh University Press. 1998.
\bibitem{RUD98} J. Rudman: The State of Authorship Attribution Studies. Some Problems and Solutions. Kluwer Academic Publishers, 1998.
\bibitem{SFK01} E. Stamatatos, N. Fakotakis, G. Kokkinakis: Computer-based Authorship Attribution Without Lexical Measures. Journal on \textit{Computers and the Humanities}, Springer, 35:193-214, 2001.
\bibitem{SE00} G. Kokkinakis, E. Stamatatos, N. Fakotakis: Automatic text categorization in terms of genre and author. Computational Linguistics, 26(4):471-495, 2000.
\bibitem{SMI83} M. W. Smith: Recent experience and new developments of methods for the determination of authorship. Association for Literary and Linguistic Computing Bulletin, 11:73-82, 1983.
\bibitem{T87} T. Mendenhall: The characteristic curves of composition. Science, pp. 214:237249. 1887.
\bibitem{TAL72} D. R. Tallentire: An appraisal of methods and models in computational stylistics, with particular reference to author attribution. PhD Thesis, Univesity of Cambridge. 1972.
\end{thebibliography}
}
\end{document}